\documentclass[final,3p,times,twocolumn]{elsarticle}
\usepackage{amssymb}
\usepackage{epstopdf}% To incorporate .eps illustrations using PDFLaTeX, etc.
\usepackage{url}
\usepackage[caption=false]{subfig}% Support for small, `sub' figures and tables
\usepackage{moreverb}
\usepackage{color}

\journal{Robotics And Autonomous Robots}

\begin{document}

\begin{frontmatter}

\title{Towards a Robotic Intrusion Prevention System: Combining Security and Safety in Cognitive Social Robots}

\author[irl]{Francisco Mart\'in}
\author[hc]{Enrique Soriano-Salvador}
\author[irl]{Jos\'e Miguel Guerrero}
\author[hc]{Gorka Guardiola M\'uzquiz}
\author[irl]{Juan Carlos Manzanares}
\author[ule]{Francisco J. Rodr\'iguez}

\affiliation[irl]{organization={Intelligent Robotics Lab, Rey Juan Carlos University},%Department and Organization
            addressline={EIF},
            city={Fuenlabrada},
            postcode={28943},
            state={Madrid},
            country={Spain}}

\affiliation[hc]{organization={HoneComp, Rey Juan Carlos University},%Department and Organization
            addressline={EIF},
            city={Fuenlabrada},
            postcode={28943},
            state={Madrid},
            country={Spain}}

\affiliation[ule]{organization={Robotics Group, Universidad de Leon},%Department and Organization
            addressline={Campus Vegazana},
            city={Leon},
            postcode={24071},
            state={Leon},
            country={Spain}}

\begin{abstract}
Social Robots need to be safe and reliable to share their space with humans. This paper reports
on the first results of a research project that aims to create more safe and reliable,
intelligent autonomous robots by investigating the implications and interactions between
cybersecurity and safety. We propose creating a robotic intrusion prevention system (RIPS) that
follows a novel approach to detect and mitigate intrusions in cognitive social robot systems
and other cyber-physical systems.
The RIPS detects threats at the robotic communication level and
enables mitigation of the cyber-physical threats by
using \emph{System Modes} to define what part of the
robotic system reduces or limits its functionality while the system is compromised. We demonstrate
the validity of our approach by applying it to a  cognitive architecture running in a real social
robot that preserves the privacy and safety of humans while facing several cyber attack situations.

\end{abstract}



\begin{keyword}
Social Robotics \sep Safety \sep Security \sep Cognitve Robotics
% %% or \MSC[2008] code \sep code (2000 is the default)
\end{keyword}

\end{frontmatter}


\section{Introduction}

Cognitive robots augment their autonomy, enabling them to deploy in increasingly open-ended
environments~\cite{Vernon:2016}\cite{Sussman:2021}. These advances offer enormous possibilities for improvements in the human economy
and well-being~\cite{FoschVillaronga2015CreationOA}. However, it also poses high risks that are difficult to assess and control by humans\cite{fosch21}.
Our group's research aims to create safer and more reliable intelligent autonomous robots by
investigating the implications and interactions between cybersecurity, safety, and explainability. Robots
are increasingly common in our society: vacuum cleaners, company robots, caregivers, and autonomous
vehicles, among others. Robots need to be safe, respectful of privacy, not manipulable by external agents,
and capable of explaining their behavior. This paper focuses on the first two components:
safety and security.

Safety has been an essential element in robotics since the first industrial robots~\cite{braganca19}\cite{dhillon2002robot}\cite{colgate2008safety}.
A typical implementation
of safety mechanisms is the use of modes of System Modes~\cite{9474424}
The presence of people in the work area of a robot changes the
operating mode to a different one that limits the damage that can be caused to people, for example by
reducing the speed or power of the motors.

Regarding security, the field  of robotics (and more so in mobile robotics)
has tended to rule out the use of secure protocols and
tools because robots have traditionally been little-exposed systems that
functioned in well-controlled research environments.
With the increasingly common deployment in organizations and domestic environments,
this assumption is no longer valid, requiring the establishment of mechanisms
that ensures that a robotic system is not compromised or that this situation is detected and mitigated.

Intrusion Detection Systems (IDS) are tools that detect unauthorized access to a computerized system. In the event of any suspicious activity, they issue alerts. Intrusion Prevention Systems (IPS) are tools that, besides detecting intrusions, also perform actions to mitigate the threats. There are two types of IPS: Network Intrusion Prevention Systems (NIPS)
and Host Intrusion Prevention Systems (HIPS).
NIPS are network devices that monitor the wireless and wired traffic and mitigate the threats by dropping suspect packets,  blocking and resetting connections, etc.  HIPS are installed in the endpoints to analyze and filter the nodes' traffic.
Those systems apply different detection techniques, such as
rule-based approaches,
signature-based analysis (binary patterns in the data),
anomaly-based analysis (detection of unexpected or abnormal behavior),
policies,
and different machine learning approaches (clustering, classification, estimation, association, prediction, statistics, etc.)~\cite{9299068,9620099,sandhu2011survey}. NIPS and HIPS are combined and integrated into unified threat management (UTM) solutions to secure conventional distributed systems.

As other authors~\cite{guerrero2017empirical}, we state that
conventional network IDS and IPS solutions~\cite{9299068,9620099,sandhu2011survey} are not suitable for autonomous robotic systems,
cognitive social robots, and other kinds of cyber-physical systems (CPS)
that comprise interacting digital, analog, physical, and human components engineered to function through integrated physics and logic~\cite{229211}.
In this scenario, traditional
IPS solutions can be used
to detect and prevent low level communication attacks,  but they
are not able to inspect the robotic communications properly
and apply safety related mitigation measures at the robotic level.
For this reason, we propose the creation of a Robotic Intrusion Prevention
System (RIPS) for ROS 2~\cite{doi:10.1126/scirobotics.abm6074} systems.

% {\color{red} [esoriano: Yo quitaba las preguntas de investigación porque no se pueden responder,
%al menos la RQ1 no se puede. Esto es poner el reject en bandeja. Me centraría en las aportaciones
%sin preguntas de investigación, no creo que sea ese tipo de paper]}

%The research presented in this paper attempts to answer two research questions:

%\begin{itemize}
%  \item \textbf{RQ1}: What kind of detection mechanisms should the RIPS provide?
%  \item \textbf{RQ2}: How does the mitigation part of the RIPS interact with a %cognitive system that implements safety with System Modes?
%\end{itemize}

%\emph{RQ1} is about what kind of mechanisms and rules have to be included in the RIPS,
%focusing on the application layer of ROS 2.
%We hypothesize that the RIPS is effective by examining the data flow and interconnections
%between the robotic application components. In the case of a cognitive architecture implemented
%in ROS 2, it is effective by examining the participants and data in a computation graph.

%\emph{RQ2} addresses how to mitigate the cyber-physical threats once an intrusion is detected.
%We hypothesize that by extending the use of modes of operation (typical in industrial robotic systems),
%it is possible to preserve people's privacy and integrity in robots deployed in environments
%shared with people.

It is common to use operating modes for safety. The ISO 13482:2014 standard, which covers
robots, spacially personal care robots, defines various modes of operation in which a robot can operate safely.
System Modes~\cite{9474424} are a way of implementing modes of operation in robots since they provide an
architectural scheme to separate the management of contingencies and modes of operation
independently of the application's logic.

In this work, we will define some modes of operation similar to those proposed in safety, but
that can represent the threat level detected by a RIPS system. We are convinced that in this
way, it is easy to integrate into any robotic system the principles of safety, privacy, and
integrity of the information without affecting the main logic of an application. , the system
is reactively reconfigured to adapt to the perceived threat level and the mitigation that can
be carried out.

The experimental validation of our approach will present a set of appropriate metrics to
validate or reject the hypotheses raised and thus answer the research questions raised.
The main contributions of this paper are:
\begin{enumerate}
    \item The analysis and design of an system to detect intrusions in a social robot.
    \item A novel approach to deploy safety in social robots using system modes.
    \item A Security/Safety mechanism that combines the previous points in the same robot system.
\end{enumerate}

This paper is structured as follows: In section \ref{sec:related}, we put our approach
in context, analyzing the existing works in this area. In section \ref{sec:arq}, we
present our cognitive architecture, which we use as reference to develop our proposal
in section \ref{sec:secsaf}. We detail out the experimental validation in section
\ref{sec:exps}, which serves to validate or reject our hypotheses. Finally, we
answer our research questions in the conclusions presented in section \ref{sec:conclusions}.

\section{Related Work}
\label{sec:related}

\subsection{Security in Robots and Autonomous Systems}

Different studies focusing on generic cybersecurity aspects
for robotic environments (e.g., threats, vulnerabilities, attacks, risks, etc.)
can be found in the literature. Cerrudo et al.~\cite{cerrudo2017hacking}
studied generic security issues
in robotics, such as insecure communications, authentication issues, authorization
issues, weak cryptography, privacy issues, etc. They also provided some
example attack scenarios for different scopes (military, industrial, etc.).
Clark et al.~\cite{clark2017cybersecurity} published a
paper describing real and potential threats to robotics
at the hardware, firmware, operating system, and application levels.
They also analyzed the impact of such attacks and presented a set of
countermeasures for each level.
Basan et al.~\cite{basan2019analysis} proposed a methodology
to analyze the security of a network of mobile robots.
They classified the structural and functional
characteristics of robotic systems and analyzed potential attackers,
their goals, and capabilities.

Three surveys on threats and attacks on robotic systems have been
published recently:
(i) Archibald et al.~\cite{archibald2017survey} published a
survey on security in robotic systems in 2017. They divided the
robotic system vulnerabilities in different classes:
physical,
sensor,
communication,
software,
system-level,
and user vulnerabilities. They also described potential physical and
logical attacks;
(ii) In 2019, Jahan et al.~\cite{jahan2019security}
published a survey on security modeling of autonomous systems.
This work reviewed the historical evolution, approaches and trends,
and analyzed the different types of autonomous systems (autonomous vehicles, IoT, swarms,
robots, etc.). They also presented a set survey papers regarding
the modeling of the systems and possible attacks and
a discussion on future research directions and challenges;
(iii) In 2021, Yaacoub et al.~\cite{yaacoub2021robotics} published  a survey on
the main security vulnerabilities, threats, risks, and their impacts,
for robotics environments.
There are also multiple recent surveys on the security of
autonomous vehicles (CAVs)~\cite{cavthreatssurvey,kim2021cybersecurity,limbasiya2022systematic,kukkala2022roadmap,burzio2018cybersecurity}.

The security of ROS and ROS 2 has been analyzed in multiple studies~\cite{mcclean2013preliminary,portugal2017role,jeong2017study,Dieber2020,yu2021data,rivera2019rosploit,abeykoon2017forensic,demarinis2019scanning,breiling2017,toris2014,dieber2016,white2016,secureros,MARTIN201895,Jongkilros2,diluoffo2018robot,fernandez2020performance,white2019network,ddsblackhat21}.
As far as we know, there is not any IPS for ROS 2 systems.
There are different tools for detect and debug incorrect programs
and monitor ROS1 and ROS 2 systems.
HAROS~\cite{9474545} is a framework used to detect incorrect
code in ROS 2 applications. The static analyzer framework
inspects the source code and creates a formal model in a specific language
designed for capturing ROS 2 properties (scope, patterns, etc.). Those properties
are used for testing.

ARNI~\cite{bihlmaier2014increasing,bihlmaier2016advanced} is a framework
to monitor ROS1 systems. It is used to find configuration errors and
bottlenecks in the communications. The framework allows the user to define
a reference state for the whole ROS1 system. Then, it continuously compares the
reference state to the actual state. If there are deviations, the user is  warned
through a dashboard. The user can also define countermeasures that are executed
by the system.

Drums~\cite{monajjemi2015drums} is a tool for monitoring and debugging
ROS1 systems. It permits the user to inspect the node graph, audit services, and communication channels as
native processes and network channels. In addition, it allows the user
to monitor host
resources (CPU, memory, etc.) and store the collected information in a database.
Drums provides a dashboard and uses a data-driven anomaly detection
software~\cite{skyline}, but it does not provide IPS capabilities.

Vulcanexus~\cite{vulcanexus} is a software stack for ROS 2 that provides
a set of libraries, tools, simulators, etc.  It includes a ROS 2 monitor
to track communications performance with the eProsima Fast DDS
implementation. It provides real-time information about the communications
(packet loss, latency, throughput,  etc.), that can be used
to detect bottlenecks and other issues.

ROS-FM~\cite{rosFM} is a network monitoring framework for ROS1
and ROS 2 systems. It is based on Berkely Packet Filters (eBPF) and eXpress Data Path
(XDP). It also includes a security policy enforcement tool and a
visualization application. The Berkely Packet Filters (eBPF)~\cite{ebpf}
is used to run sandboxed programs in the operating system's kernel.
Those programs are compiled
to  byte-code, which is executed by a BFP virtual machine inside the kernel.
eBPF is used to add additional
capabilities to the operating system at runtime, usually for
packet filtering. The eXpress Data Path (XDP)~\cite{hoiland2018express}
is a programmable layer in the operating system network stack.
The ROS-FM visualization application is based on Vector~\cite{netflixvector}.
Those components form a high-performance monitoring tool for
ROS1 and ROS 2. Nevertheless, ROS-FM does not provide mechanisms
to detect anomalous ROS 2 traffic or trigger actions to mitigate threats in the
robotic system.

ROS-defender~\cite{rivera2019ros} integrates three different tools: a Security
Event Management System (SIEM), an anomaly detection system (ROSWatch)
plus a intrusion prevention system (ROSDN),
and a firewall for ROS1. They are built upon a Software Deﬁned
Networking (SDN) framework (OpenFlow v1.3).
ROSWatch inspects the network and the logs in order to detect intrusions.
It uses a pattern matching model to detect attacks in data flows and logs.
The rules use identifiers such as nodes and topics and
can be classified into priority
classes.
ROSDN uses SDN to replace the standard ROS network communication.
This way, it is able to filter the communications and ban compromised
ROS1 nodes from the network.
The main problem of this approach is the dependency on SDN technology.
In addition, the performance of this system
is criticized by its own authors~\cite{rosFM}.
ROS-defender is not available for ROS 2. In addition, this system
focuses on the conventional IPS procedures (message filtering, etc.).

Different methods have been proposed to detect attacks in robotic environments.
Urbina et al.~\cite{10.1145/2976749.2978388}
presented a survey on attack detection methods in industrial control
systems (ICSs).
Vuong et al.~\cite{7363359} presented a method to detect attacks
by using the data collected from on-board systems and processes.
They used a decision tree-based method for detecting attacks
using both cyber and physical features that can be measured by its on-board
systems and processes.
Guerrero-Higueras et al.~\cite{guerrero2017empirical}
conducted an empirical analysis to demonstrate that there are
statistically meaningful differences in the data provided by
beacon-based Real Time Location Systems (RTLSs) when the robot is under
attack. They also described some precedents for detecting attacks through
sensor data, by using statistical and Machine Learning techniques.
Sabaliauskaite et al.~\cite{7423162}
proposed a method based on a statistical technique named
CUSUM~\cite{cardenas2011attacks}
to detect stealthy attacks that are designed
to avoid detection by using knowledge of the system’s
model.

\subsection{Safety in Robotics}

Safety is defined as the condition of being protected from or unlikely to cause danger,
risk, or injury. It is a critical issue in Industrial Robotics, where more than 1.5 million
industrial robots are operating in factories worldwide. However, its introduction created
other problems, mostly in those Human-Robot collaborative scenarios. One of the main objectives
of deploying robots in factories was to prevent the operators engage in hazardous tasks or
environments~\cite{dhillon2002robot}.

A complete collection of safety standards has been developed in this field to reduce the risk
of accidents when robots interact with humans. The concepts of robot design, manufacture,
installation, operational processes, maintenance and decommission are key when we are planned
to deploy a robot in public spaces and standards such as UNE-EN ISO 10218:1 proposes the rules
for Industrial Robots. Three different iterations 2006, 2008 until the current 10218:1 2012 ISO
covers all type of risks associated to non planned tasks such as start-ups or human interaction
as well as definitions for regular operation.

This research goes outside industrial robot, consequently, it is necessary to explore alternative
standards such as ISO 13482:2014 which cover robots and robotic devices, particularly safety
requirements for personal care robots. This standard does not cover robots moving faster than
20 km/h, robot toys, water-borne and flying robots, military robots or medical devices such as
DaVinci robot. This standard do define four different Operational Modes for care robots:
the Autonomous mode; Semi-autonomous mode; Manual mode Maintenance mode.
These modes should not be confused with the system modes proposed in this research here, but should be used
as a generalization of robots deployed in public spaces.

Previous standards do need to cover current Human-Robot Interaction scenarios~\cite{colgate2008safety} and
it is necessary to explore collaborative safety features presented in research
literature~\cite{vicentini2020terminology}. Thus, it is necessary to do a detailed overview in the field
and review current research works in the field as the works proposed in working group IEEE7009~\cite{farrell2021evolution}.
Privacy is considered a recommendation \cite{MITKA20121888}, although the concerns on this matter are nowadays beyond all
doubt.

There are no references in the scientific literature about implementable models for all
these issues for robotic systems. Only models for physical safety can be found. For
instance, in \cite{FoschVillaronga2015CreationOA} the model Care Robot Impact Assessment
(CRIA), based on the ISO 13482:2014, is proposed. This standard is devoted to safety
in physical interaction between humans and robots. The European Task Force \cite{RePEc:eps:cepswp:26204}
on Artificial Intelligence spent most of its recommendations on the need for
this modelization, from the discovery phase, using honeypots for the cybersecurity
dimension, to the accuracy, soundness, and reproducibility of the systems based on
machine learning, that is, an explainability model that takes also into account the
possible cybersecurity implications. In another work~\cite{10.3389/frobt.2022.1002226} a
safety framework based on operational modes is proposed to deal with human-robot interaction.

The modes of operation are used in robotics for safety and error handling. Modeling
languages such as AADL \cite{FeilerDependabilityModeling2007} or MARTE \cite{quadri:inria-00486845} consider modes of
operation one of the primary mechanisms to address this issue, and they have been
successfully used in safety in several works \cite{jsan10030048}\cite{4630122}.
Safety cares about the possible damage a robot may cause in its environment, while
security aims at ensuring that the domain does not disturb the robot’s operation \cite{DBLP:journals/corr/abs-1806-06681}.

\subsection{Safety and Security}

Lera et al.~\cite{lera2017cybersecurity} published
a study on attacks associated with service robots, presenting
a risk taxonomy for this kind of robotic applications. This work distinguises
between security and safety threats.

In \cite{akalin19,akalin19b}, the authors examined how feedback given by a robot agent influences the various facets of participant experience in robot-assisted training: robot acceptance, sense of safety and security, attitude towards robots and task performance. The results suggested that the robot with flattering and positive feedback was appreciated by older people in general, even if the feedback did not necessarily correspond to objective measures such as performance. Participants in these groups felt better about the interaction and the robot.

In \cite{fosch21} addresses the interplay between robots, cybersecurity, and safety from a European legal perspective, a topic under-explored by current technical and legal literature. The authors illustrate cybersecurity challenges and their subsequent safety implications with the concrete example of care robots.

In \cite{tadokoro19} shows an overview of the ImPACT Tough Robotics Challenge and Strategy for Disruptive Innovation in Safety and Security (ImPACT-TRC), which is a national project of the Japan Cabinet Office  that focuses on tough robotic technologies to provide solutions to disaster response, recovery, and preparedness.

Another brief overview, the so-called Industry 4.0, is shown in \cite{braganca19, komenda19}, where authors describe how collaborative robots can be used to support human workers in Industry 4.0 manufacturing environments and how human-robot interaction might also have some risks if human factors considerations are not taken into account in the process.

\section{Cognitive Architecture for Social Robots}
\label{sec:arq}

Moving the robots from industrial cages to public spaces arises interesting research questions surrounding the field of social and assistive robots. When a robot faces a daily life task in a supermarket or in a day care center it has to accomplish new goals. These scenarios present  non-stationary
elements that have interaction with it and the humans surrounding in non-repetitive manner. Thus, it becomes necessary to define and bound the the set of constraints at decision making level for managing long-term task planning and immediate behavioral approaches.

To achieve that, it is necessary  to build and implement those models of cognition that reassemble some aspects of mind as such, knowledge, motivations, generality, and completeness. Thus, cognitive architectures ~\cite{Lieto:2018} from an artificial intelligence should provide not only the mechanisms to develop robot behaviors but also how each decision affects their actions and behavior as well as those individuals in their surrounding.

Vernon\cite{Vernon:2016}'s research in RockEU2 presents the eleven  functional abilities that a robot should achieve successfully in order to provide those cognitive process that will allow it.
However, roboticist researchers~\cite{Sussman:2021} usually present a cognitive architecture as the technical mechanisms that provides the autonomy to set long-term and short-term intermediate goals to achieve those tasks required by users or robot needs.

%ROBOT

The functional abilities of the robot and the pipeline to accomplish
tasks is solved in various different ways. The classic one is the sense-plan-act approach.
In this approach, a high level system is providing decision making
using knowledge declared in formalized descriptive files and afterwards
applying it to general-purpose planning systems. ROS 2 Planning System (aka Plansys2)~\cite{martin2021plansys2} is a ROS 2 solution whose aim is to provide a reliable, simple, and efficient PDDL-based planning system
for robots.

For more specific action scheduling, behavior trees (BTs) have been adapted from software video-games scenarios for its use in robotics. Different researchers \cite{Colledanchise:2016,French:2019} present them as a generalization for
different classic control approaches in robotics such as hybrid control Systems, finite state machines, generalize sequential behavior compositions, subsumption architecture and decision trees. Thus, current behavior generation is deployed using BTs.

\begin{figure}[ht!]
\centering
   \includegraphics[width=0.5\textwidth]{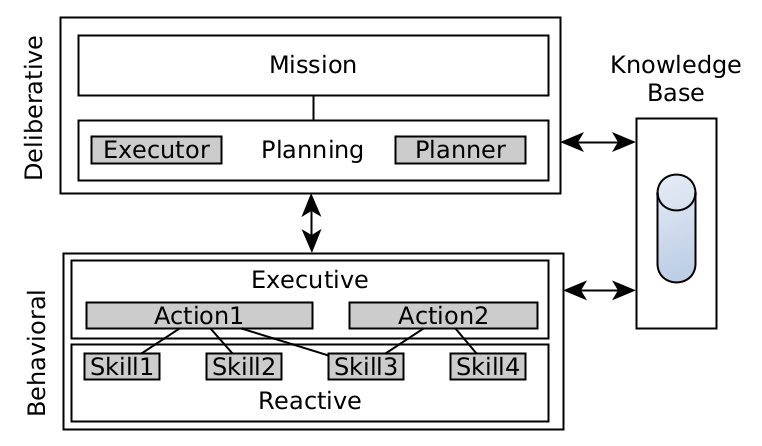}
    \caption{Funcional components of MERLIN2~\cite{gonzalez2023MERLIN2}.} \label{fig:Merlin2}
\end{figure}

Both, planning and behavioral approaches are usually presented in a single architecture. It is common to articulate a hybrid architecture for controlling and managing the behavior of a robot.
Examples such as LAAIR~\cite{jiang2018laair}, Aura~\cite{arkin1997aura}, or MERLIN2~\cite{gonzalez2023MERLIN2} illustrated in Figure~\ref{fig:Merlin2}, are composed of a deliberative system, which is responsible for making decisions and creating plans, and a behavioral system, which has the actions and skills that the robot can take.
These systems consist of layers that are built on top of each other. The deliberative system is composed of a Mission layer and a Planning layer. The behavioral system is composed of the Executive layer (robot actions) and the Reactive layer (robot skills such as Nav2 or speech generation).

\section{Intrusion Detection in ROS 2 Systems}
\label{sec:rips}

This section is a formal abstract description of the set of features that
can meet the minimum requirements of a RIPS for ROS 2 systems.
ROS 2 is based on the DDS (Data Distribution Service)
middleware~\cite{dds}. ROS 2 components,
or nodes, usually communicate following the publisher/subscriber model
by using topics\footnote{ROS 2 also provides other
mechanisms for client/server interactions
(i.e., \emph{services}).}. We will focus on this communication
model.

The RIPS will run as a ROS 2 node that monitors the interactions of
the rest of nodes. It includes an engine that
evaluates rules defined by the user.
Those rules tell RIPS how to process the ROS 2 traffic and how to react
for each monitored message.

Basically, a rule is composed by these parts:

\begin{itemize}
    \item Name: The identifier of the rule, as a string of characters.
    \item Expression: A boolean expression.
    \item Actions: A set of actions that will be executed when the boolean
    expression evaluates to \texttt{true}.
\end{itemize}

The abstract types used by the rules are:

\begin{itemize}
    \item \texttt{integer}
    \item \texttt{floating point number}
    \item \texttt{string of characters}
    \item \texttt{boolean}
    \item \texttt{set of \emph{basic-type}}
\end{itemize}

\subsection{Alert levels}

The RIPS defines an ordered set of \emph{levels}, each with a string
identifier. The alert level describes the current security state of the system
(i.e., how compromised may be the robotic system). This will define the
set of permitted actions for the robot (i.e., the privilege limiting which
operations can be performed).

The system always starts at a defined level. By default, it only can
transition to a more restrictive (lower privileged) level.
Conversely, the alert level cannot be decreased autonomously by default,
it requires human intervention. Nevertheless, the user can
define explicit transitions to lower restrictive levels by using a
keyword (\texttt{soft}). For example, we can define four levels:

\begin{itemize}
    \item \texttt{DEFAULT}
    \item \texttt{soft ALERT}
    \item \texttt{COMPROMISED}
    \item \texttt{HALT}
\end{itemize}

In this case, if the current system level is \texttt{ALERT}, it can de-escalate
and switch to the \texttt{DEFAULT} level autonomously. On the other hand,
if the system is in the \texttt{COMPROMISED}
level, it can not switch to the \texttt{ALERT} level autonomously. That requires
human intervention.

When the current alert level is changed by the RIPS,
two procedures are executed: The one needed to leave the current
level and the one needed to enter the new level. This mechanism follows the traditional Unix runlevel approach. This mechanism is flexible: In the case of using System Modes (the proposal of this work), the alert levels can be directly mapped to System Modes. In another case, it permits the definition of generic procedures (e.g., Python or shell scripts) to launch the selected countermeasures or
mitigation actions. Both approaches can be combined together.

For example, suppose that the current alert level is \texttt{DEFAULT} and
the RIPS changes it to the \texttt{ALERT} level.
Then, two  procedures are executed: The one
to leave the \texttt{DEFAULT} level and the one to enter the
\texttt{ALERT} level. Then, these procedures
order the System Mode framework to change the robot's System Mode
from the \texttt{DEFAULT} mode to the \texttt{ALERT} mode.

\subsection{Predefined Global Variables}

The RIPS maintains a set of global predefined variables.
Those variables can be used in the rules' expressions.
For example, the predefined global variables include:

\begin{itemize}
    \item \texttt{Time}: Integer with the current time.
    \item \texttt{Level}: String with the current level.
    \item \texttt{Uptime}: Integer with the uptime of the system.
    \item \ldots
\end{itemize}

\subsection{Actions}

Rules are reactive to \emph{events}. Each time an \emph{event} is detected
by the RIPS, the related rules are evaluated. If a rule's expression
evaluates to \texttt{true}, its actions will be executed.
There are three classes of events:

\begin{itemize}
    \item \emph{Message}: Messages delivered to a topic.
    \item \emph{Graph}: Changes in the ROS 2 computation graph.
    \item \emph{External}: External events.
\end{itemize}

We identify four types of actions:

\begin{itemize}
    \item \texttt{alert}: An alert event is generated and logged by the system.

    \item \texttt{set}: This action defines or modifies
    variables of the system. The user can define new variables to maintain
    the context of the rules. Simple arithmetic or string expressions are allowed.
    For example, the user can define a counter which is incremented
    each time some specific rules evaluate to \texttt{true}. This variable,
    together with the current time, can be used to write a rule
    for detecting flooding attacks.

    \item \texttt{exec}: This action just executes an external program
    with arguments. For example, the rule can execute
    a simple Python script to configure a new filter in a firewall
    in order to drop some types of messages.

    \item \texttt{trigger}: This action changes the current alert level,
    triggering a transition. The procedure to exit the current alert level
    and the procedure to enter the new level is executed.
    It has one argument, the new alert level (that must be a lower privileged
    level if the current level is not marked as \emph{soft}).
\end{itemize}

A rule can execute more than one action sequentially. Actions are executed
in chains.
In a chain, the result of each action is evaluated to consider if the next
action is executed depending on the operator chaining them.
These operators can
be used to execute actions depending on the the result of the previous action in the chain:

\begin{itemize}
    \item $\rightarrow$ executes the right action only if the left action
    is successful.
    \item $\not\rightarrow$ executes the right action only if the left action
    is not successful.
    \item $,$  executes the right action after the left action without considering
    if it is successful or not (but the chain of actions has to be executed up to that
    point to even consider the execution).
    \item $\square$ terminates the chain of actions.
\end{itemize}

More than one chain can be executed when a rule is triggered.

For example, a rule R can execute the following three chains of actions:

{\scriptsize
\begin{center}
$alert(``info: rule\ R\ activated")\square$\\
$exec(usb\_alarm) \not\rightarrow  alert(``warning: usb\_alarm\ failed")\square$\\
$trigger(HALT) \square$
\end{center}
}

Therefore:

\begin{enumerate}
    \item A message is logged in the system to inform of the rule activation.
    \item An external program named  \texttt{usb\_alarm} is executed. This program commands the system to activate
        a physical alarm (e.g. a USB alarm with sound and lights) connected to the computer.
    \item If the program \texttt{usb\_alarm} fails, a new warning message about the failure is
        logged in the system. Else, that warning message is not logged.
    \item The \texttt{HALT} alert level is triggered and the corresponding procedures are executed.
\end{enumerate}

\subsection{Expressions}

Expressions should not have lateral effects. Note that, even though an external
the program can be executed as a consequence of the evaluation of a subexpression, the
evaluation of the whole expression should be idempotent.

In this abstract description,
we represent the subexpressions as boolean functions.
The evaluation precedence is from right to left.
The boolean operators \texttt{and}, \texttt{or}, and \texttt{not} are
available to create the expression. It applies a shortcut evaluation of
such expressions.

Subexpressions of different classes of events cannot be combined.
Thus, the rule has an implicit \emph{type} (i.e., \emph{message},
\emph{graph} or \emph{external}), which is inferred from its expression.

We consider that the following set of subexpressions
provide enough expressivity to properly
model the rules to detect threats in the ROS 2
publisher/subscriber communication scheme.

\subsubsection{Subexpressions for message events \label{msgsubexpr}}

\begin{itemize}

%% security MUST be enabled for RIPS, it makes no sense to run RIPS
%% without it
%%    \item \texttt{secure()}: This function returns \texttt{true} if the message is
%%    secured by ROS 2 security mechanisms (authentication, confidentiality, access control,
%%    etc.).

%% RIPS will run only on domain 0. Other domains should be forbidden.
%%    \item \texttt{domain(d:int)}: This function returns \texttt{true} if the DDS domain
%%    is the specified in the parameter.

    \item \texttt{topicin(topics:set of string)}: This function returns \texttt{true} if
    the topic name of the message is included in the specified set of topic names.

    \item \texttt{topicmatches(regex:string)}: The function's parameter is a regular
    expression. The function returns \texttt{true} if the topic name (the whole name,
    including the namespace) of a message matches the regular expression.

    \item \texttt{publishercount(min:int, max:int)}: The function returns \texttt{true}
    if the number of participants
    publishing to the topic of this message are in the range defined by the interval
    $[min..max]$.

    \item \texttt{subscribercount(min:int, max:int)}: The function returns \texttt{true}
    if the number of participants
    subscribed to the topic of this message are in the range defined by the interval $[min..max]$.

    \item  \texttt{publishersinclude(pubs:set of string)}: The parameter is a set of strings with
    the names of participants (nodes). The function returns \texttt{true} if all  the
    participants of the set are publishers for the topic of the message.

    \item  \texttt{subscribersinclude(subs:set of string)}: The parameter is a set of strings with
    the names of participants (nodes). The function returns \texttt{true} if all  the
    participants of the set are subscribers of the topic of the message.

    \item  \texttt{publishers(pubs: set of string)}: The parameter is a set of strings with
    the names of participants (nodes). The function returns \texttt{true} if the
    participants of the set are exactly the publishers for the topic of the message.

    \item  \texttt{subscribers(subs: set of string)}: The parameter is a set of strings with
    the names of participants (nodes). The function returns \texttt{true} if the
    participants of the array are exactly the subscribers of the topic of the message.

    \item \texttt{msgtypein(msgs: set of string)}: The parameter is a set of strings with
    the names of ROS 2 message types (\texttt{"std\_msgs"}, \texttt{"sensor\_msgs"},
    \texttt{"diagnostic\_msgs"}, \texttt{"geometry\_msgs"}, \texttt{"nav\_msgs"},
    \texttt{"shape\_msgs"}, \texttt{"stereo\_msgs"}, \texttt{"trajectory\_msgs"}, and
    \texttt{"visualization\_msgs"}). The function returns \texttt{true} if the
    message type of the message is included in the set.

    \item  \texttt{msgsubtype(msg:string, submsg:string)}: The first parameter is the
    type of the message, the second one is the subtype of the message.
    For example, if the type is \texttt{"std\_msgs"},
    the subtype can be \texttt{"ColorRGBA"},
    \texttt{"Empty"}, or \texttt{"Header"}. The function returns \texttt{true} if
    the message has the specified type and subtype.

    \item \texttt{plugin(id:string)}: The plugin specified by the parameter
    is invoked. A plugin is a program that is
    able to inspect and analyze the ROS 2 message.
    Plugins permit the user to create extensions
    for custom analysis. For example, the user can create a plugin
    that analyzes the frames sent by a camera component in order
    to detect a \emph{camera blinding} attack~\cite{petit2015remote,yan2016can}.

    \item \texttt{payload(path:string)}: This function invokes YARA to find
    patterns in the payload of the message. YARA~\cite{yara} is the de facto
    standard for textual and binary pattern matching in malware analysis.
    It is widely used to detect malware \emph{signatures}.
    The RIPS must be able to inspect the payload of the robotic messages, because
    they can be used to inject malicious code in the components.
    This function executes the YARA engine over the
    payload of the message to find the patterns specified by
    the YARA rules stored in the file passed as argument.
    If any of these YARA rules matches, the function returns \texttt{true}.

    \item \texttt{eval(var:string, operator:string, value:string)}
    This function is used to check the values of the variables.
    The second parameter is the textual representation of an operator.
    The following operators are defined for all the basic types:
    $==, !=, <, >, <=, >=$.
    The third parameter is the textual representation of the value for the
    corresponding basic type.
\end{itemize}

\subsubsection{Subexpressions for graph events}

The ROS 2 computation graph is defined as the network elements processing
data together at one time.
It includes the nodes and their connections (topics, services, actions, or
parameters). The following subexpressions can be used for this kind
of event:

\begin{itemize}
    \item  \texttt{nodes(n: set of string)}: This function returns
    \texttt{true} if the current nodes of the graph
    are just the ones defined by the set
    passed as argument.

    \item  \texttt{nodesinclude(n: set of string)}:  This function returns
    \texttt{true} if the current nodes of the graph
    are included in the set passed as argument.

    \item  \texttt{nodecount(min:int, max:int)}: If the current number of
    nodes is in the interval $[min..max]$, the function returns \texttt{true}.

    \item \texttt{eval(varname:string, operator:string, value:string)}: Its
    the same function explained in Section~\ref{msgsubexpr}, used to check
    the values of variables.
\end{itemize}

The semantics of the following functions are similar to the semantics of
the previous ones (i.e., \texttt{nodes},
\texttt{nodeinclude}, and \texttt{nodecount}), but they are oriented to
topics, services, topic's subscribers, and topic's publishers:

\begin{itemize}
    \item  \texttt{topics(nodes:set of string)}
    \item  \texttt{topicsinclude(nondes:set of string)}
    \item  \texttt{topiccount(min:int, max:int)}

    \item  \texttt{services(node string, s:set of string)}
    \item  \texttt{servicesinclude(node string, s:set of string)}
    \item  \texttt{servicecount(node string, min:int, max:int)}

    \item  \texttt{topicsubscribers(topic:string, nodes:set of string)}
    \item  \texttt{topicsubscribersinclude(topic:string, nodes:set of string)}
    \item  \texttt{topicsubscribercount(topic:string, min:int, max:int)}

    \item  \texttt{topicpublishers(topic:string, nodes:set of string)}
    \item  \texttt{topicpublishersinclude(topic:string, nodes:set of string)}
    \item  \texttt{topicpublishercount(topic:string, min:int, max:int)}
\end{itemize}

\subsubsection{Subexpressions for external events}

The following subexpressions cannot be combined in a rule:

\begin{itemize}
    \item \texttt{idsalert(alert:string)}:
    The RIPS can use a conventional IDS/NIPS/HIPS to detect low-level
    network threats (e.g., to detect suspect messages at network or
    transport levels). A special expression can be used to react to a
    low level detection:
    This function returns \texttt{true}
    when the alert specified in the parameter is triggered by
    the underlying detection system.
    In our prototype, we use Snort~\cite{snort99}
    as a low-level IDS.

    \item \texttt{signal(sig:string)}:
    It can be used to react to user-defined
    Unix signals. It may be useful to trigger emergency actions from a system
    shell if necessary.
    This function returns \texttt{true} when the RIPS process
    receives a \texttt{SIGUSR1} or \texttt{SIGUSR2} signal.
\end{itemize}

\section{Implementing Safety with System Modes}
\label{sec:secsaf}

The system modes framework used in our work~\cite{9474424} allows the separation of
the system runtime configuration and the system error and contingency diagnosis,
reducing the effort for the application developer to design and implement the task,
contingency, and error handling.

\begin{figure}[h!]
  \centering
  \includegraphics[width=\linewidth]{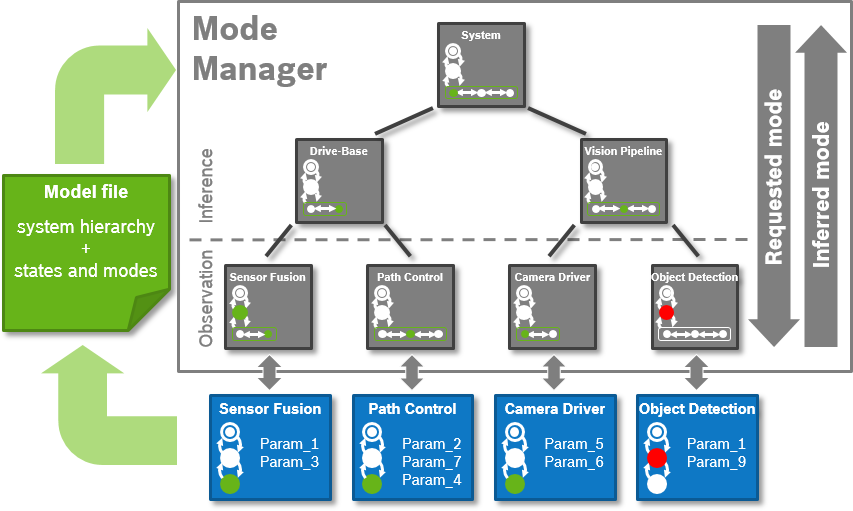}
\caption{Mode inference and mode management based on the system modes
and hierarchy file (SMH) (image from \cite{9474424}).}
  \label{fig:systemmodes}
\end{figure}

Figure \ref{fig:systemmodes} shows a system's organization in different
subsystems encompassing the components of the robotic system. A top-down flow requests
an operation mode, while a down-top flow allows it to be inferred based on configurable
rules that observe the real situation of the system. Configuration is simple, specifying
in a model file:
\begin{enumerate}
  \item The structure of systems and subsystems.
  \item Which components are active in each subsystem.
  \item The value of the parameters for each mode in each of the components.
\end{enumerate}

This framework assumes that the components that make up the cognitive architecture can
be activated and deactivated during their operation. In addition, they assume that the
operation of each of them depends on a series of configuration parameters that can be
changed and taken into account at runtime.

\begin{figure}[h!]
  \centering
  \includegraphics[width=\linewidth]{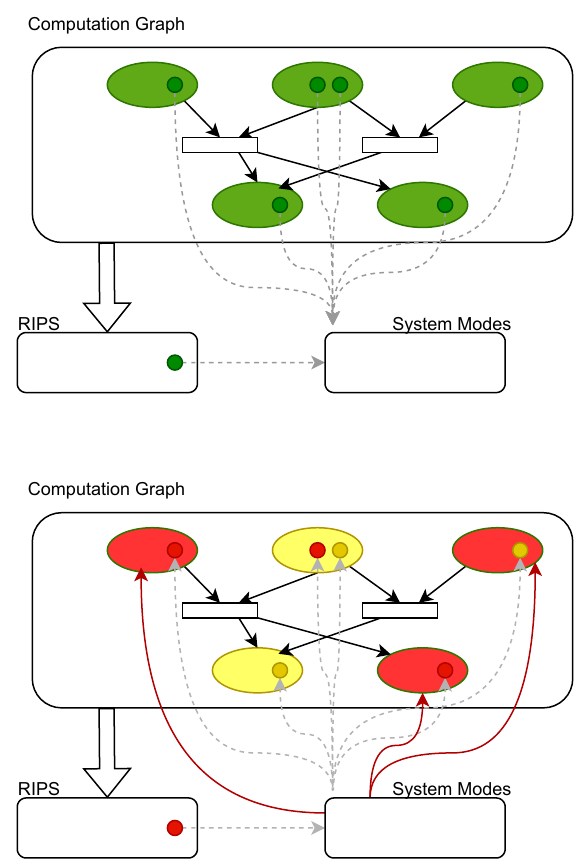}
\caption{Representation of a robotic application (a computation graph) controlled RIPS and System Modes. Small circles represent parameters with a value represented by a color. Nodes in the computation graph (ellipses) have an operation mode represented by their color. Rectangles are topics to which some nodes publish and others subscribe. The upper part represents a non compromised system (i.e. the \texttt{DEFAULT} alert level in the RIPS) and the bottom part a compromised system (i.e. the \texttt{COMPROMISED} alert level in the RIPS).}
  \label{fig:rips_sm}
\end{figure}

Suppose the scenario shown in Figure \ref{fig:rips_sm}. As explained before,
a ROS 2 application is a computation graph. The nodes (ellipses of the computation graph) in this graph are the functional elements that encode the application logic. The nodes have parameters (small circles in the figure) that modulate their operation. These nodes can be active or inactive (green is active, red is inactive, and yellow is reduced operation.
During its operation, the operating modes monitor each of the system parameters. If there is a variation in the parameters' value or the nodes' state, the System Modes infer the system's operating mode and establish which nodes should be active in the new mode and their parameters' value.
%% RIPS no mantiene ningún parámetro de ROS2 para notificar un cambio en los sistem modes.
%% Está pensado para ejecutar un script que indique a los system modes un cambio de nivel.
As explained in Section~\ref{sec:rips}, the RIPS watches the computation graph and the messages sent by the application's components. Depending on the defined rules, RIPS sets its current alert level. In this scenario, RIPS only has defined four alert levels: \texttt{DEFAULT}, \texttt{ALERT}, \texttt{COMPROMISED}, and \texttt{HALT}. While the level is \texttt{\texttt{DEFAULT}}, all nodes are up and running with their parameters in normal mode (they are green). When RIPS detects some threats (i.e. some RIPS rules are activated and the corresponding actions are performed), it changes its level to \texttt{COMPROMISED}. Then, System Modes reacts to this change and infers a new mode, in which it deactivates some nodes (red arrow and red color on nodes) and sets values for node parameters, causing some to go into a shutdown mode and others to a reduced operation (yellow).

This is an example configuration for the System Operations:

\scriptsize
%\begin{multicols}{1}
\begin{verbatimtab}
safety:
  ros__parameters:
    type: system
    parts:
      image_1_to_2
      imu_1_to_2
      odom_1_to_2
      pc2_1_to_2
      scan_1_to_2
      tf_1_to_2
      tf_static_1_to_2
      twist_2_to_1
    modes:
      __DEFAULT__:
        image_1_to_2: active
        imu_1_to_2: active
        odom_1_to_2: active
        pc2_1_to_2: active
        scan_1_to_2: active
        tf_1_to_2: active
        tf_static_1_to_2: active
        twist_2_to_1: active
        planner_server: active
        filter_mask_server: inactive
        costmap_filter_info_server: inactive
        filter_mask_server_clean: active
        costmap_filter_clean: active
      ALERT:
        image_1_to_2: inactive
        imu_1_to_2: active
        odom_1_to_2: active
        pc2_1_to_2: active
        scan_1_to_2: active
        tf_1_to_2: active
        tf_static_1_to_2: active
        twist_2_to_1: active
        planner_server: active
        filter_mask_server: active
        costmap_filter_info_server: active
        filter_mask_server_clean: inactive
        costmap_filter_clean: inactive
      COMPROMISED:
        image_1_to_2: inactive
        imu_1_to_2: active
        odom_1_to_2: active
        pc2_1_to_2: inactive
        scan_1_to_2: inactive
        tf_1_to_2: active
        tf_static_1_to_2: active
        twist_2_to_1: inactive
        planner_server: active
        filter_mask_server: active
        costmap_filter_info_server: active
        filter_mask_server_clean: inactive
        costmap_filter_clean: inactive
      HALT:
        image_1_to_2: inactive
        imu_1_to_2: inactive
        odom_1_to_2: inactive
        pc2_1_to_2: inactive
        scan_1_to_2: inactive
        tf_1_to_2: inactive
        tf_static_1_to_2: inactive
        twist_2_to_1: inactive
        planner_server: active
        filter_mask_server: inactive
        costmap_filter_info_server: inactive
        filter_mask_server_clean: inactive
        costmap_filter_clean: inactive


\end{verbatimtab}
%\end{multicols}
\normalsize

\section{Experimental Validation}
\label{sec:exps}

In this section, we experimentally validate the contributions described in this work, focused on the processes involved in security and safety in a real mobile robot. The objective is to show that security issues can be satisfied or enhanced with systems that detect and mitigate intrusion through system modes.

\begin{figure}[h!]
  \centering
  \includegraphics[height=0.7\linewidth]{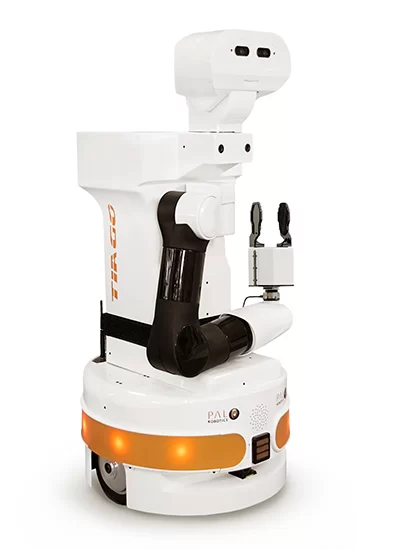}
  \includegraphics[height=0.5\linewidth]{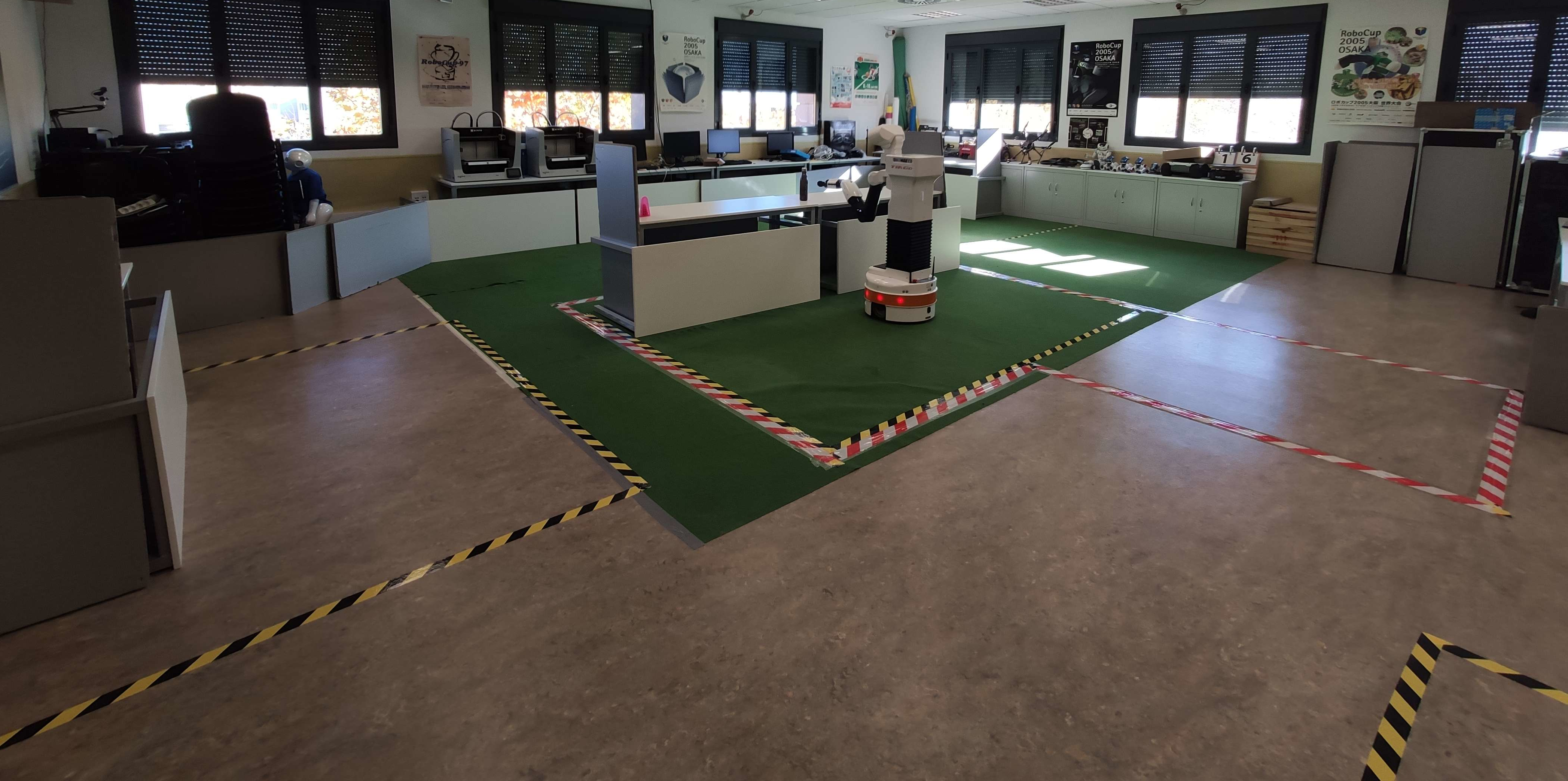}
\caption{Robot Tiago (upper) and experimental validation setup (bottom).}
  \label{fig:setup}
\end{figure}

We have validated our contribution implementing onboard a Tiago\footnote{https://pal-robotics.com/es/robots/tiago/} robot (left side of Figure \ref{fig:setup}) the cognitive architecture described in section \ref{sec:arq} that uses the system modes described in section \ref{sec:secsaf}. In detail, the Tiago robot uses Ubuntu 18.04, ROS1 Melodic, and CycloneDDS. A laptop mounted on top uses Ubuntu 22.04, ROS 2 Humble, and CycloneDDS to command the robot and send the orders simulating the intrusion. To communicate ROS1 and ROS 2, we use bridges available at (\url{https://github.com/fmrico/ros1_bridge}). The RIPS core described in Section \ref{sec:rips} is emulated in this experiment.

To demonstrate the system in operation, we have devised a scenario on the right-hand side of Figure \ref{fig:setup}, which emulates an industrial work environment, with three work zones on the edges marked with yellow-black tapes and a central high-security zone marked with red-white tape. The robot must move between these zones to perform specific tasks. If the robot detects an intrusion and is compromised, it should not circulate in the high-security zone, nor should the camera be turned on to prevent sensitive data theft.

To verify the correct functioning of the system, we have shown the evolution over time of two metrics:
\begin{enumerate}
    \item \textbf{Sending images}: We will show the hertz at which the robot driver publishes images to a topic accessible to the rest of the system and hypothetical intruders.
    \item \textbf{Navigable area}: This represents that the central area is not available to be used to navigate through it by being marked as a "forbidden" o "keep-out zone." The map initially consists of 39720 occupied cells and 1662 free ones. Detecting obstacles or activating the keep-out zone reduces free cells and increases occupied cells.
\end{enumerate}

\begin{figure}[h!]
\centering
    \includegraphics[width=0.48\linewidth]{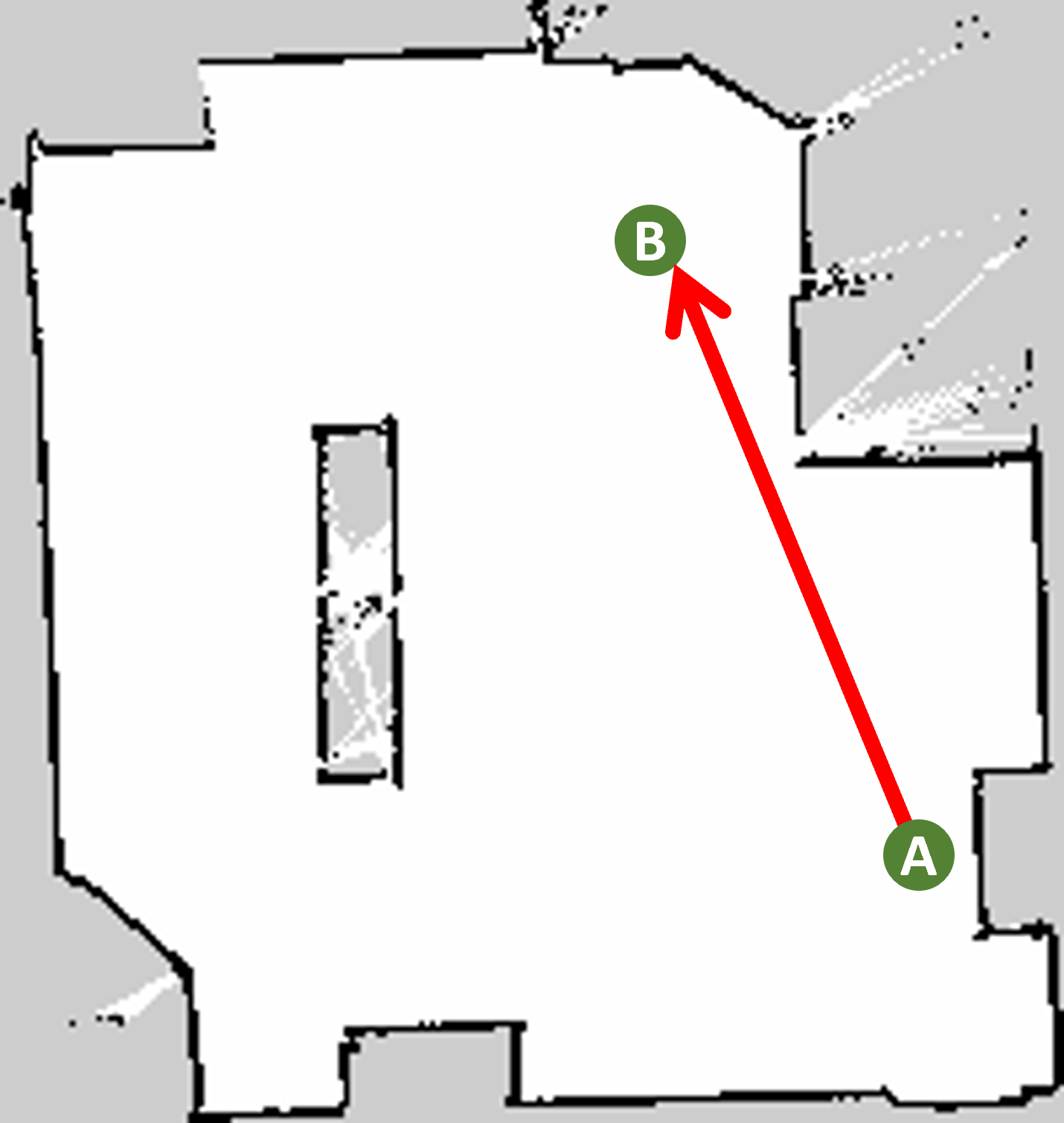}
    \caption{Initial map with start (A) and goal (B) points, and first calculated path.}
\label{fig:map1}
\end{figure}

\begin{figure}[h!]
\centering
    \includegraphics[width=0.48\linewidth]{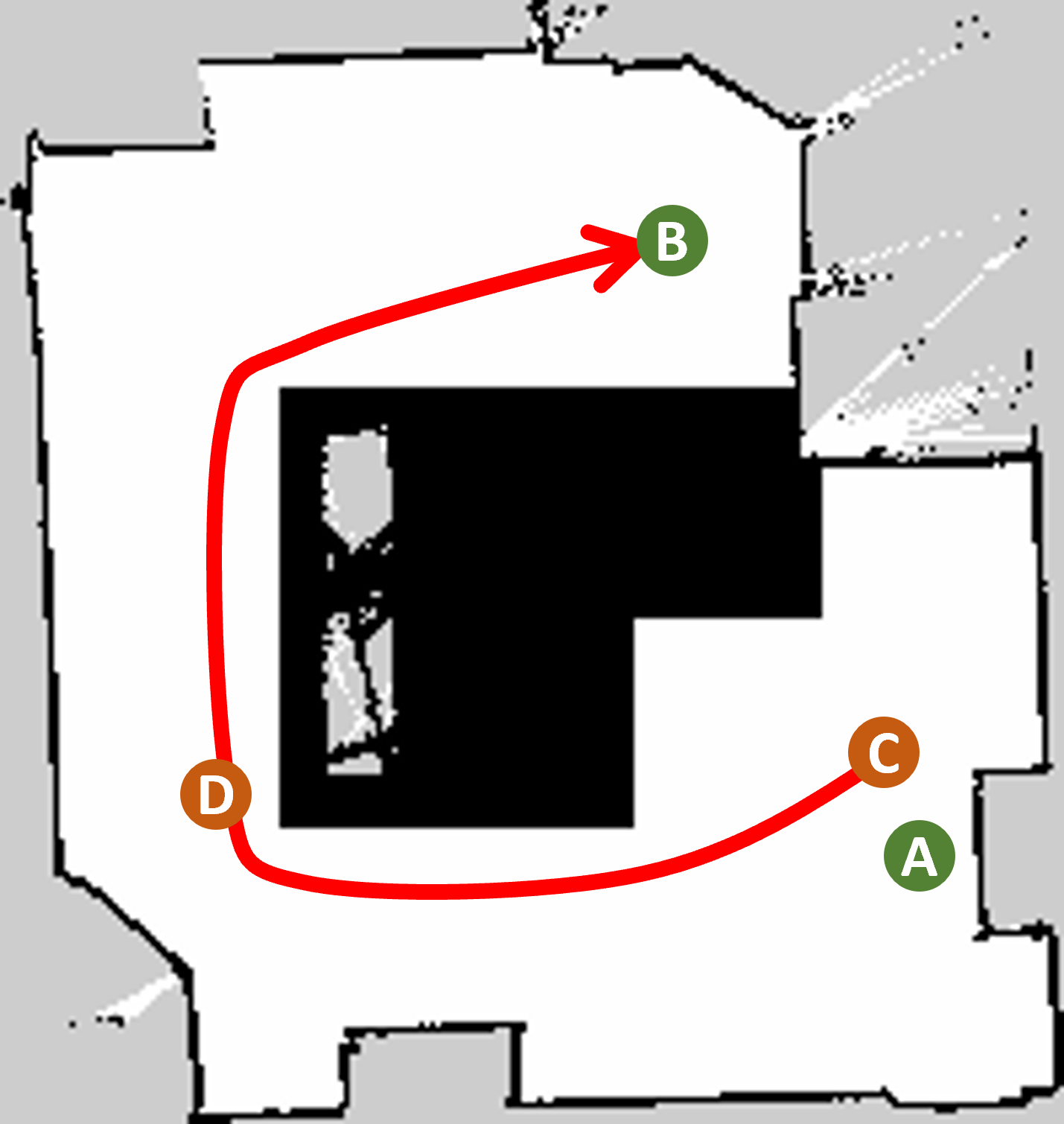}
\caption{Intrusion points (C and D), and new path.}

\label{fig:map2}
\end{figure}

\begin{figure}[h!]
\centering
    \includegraphics[width=0.48\linewidth]{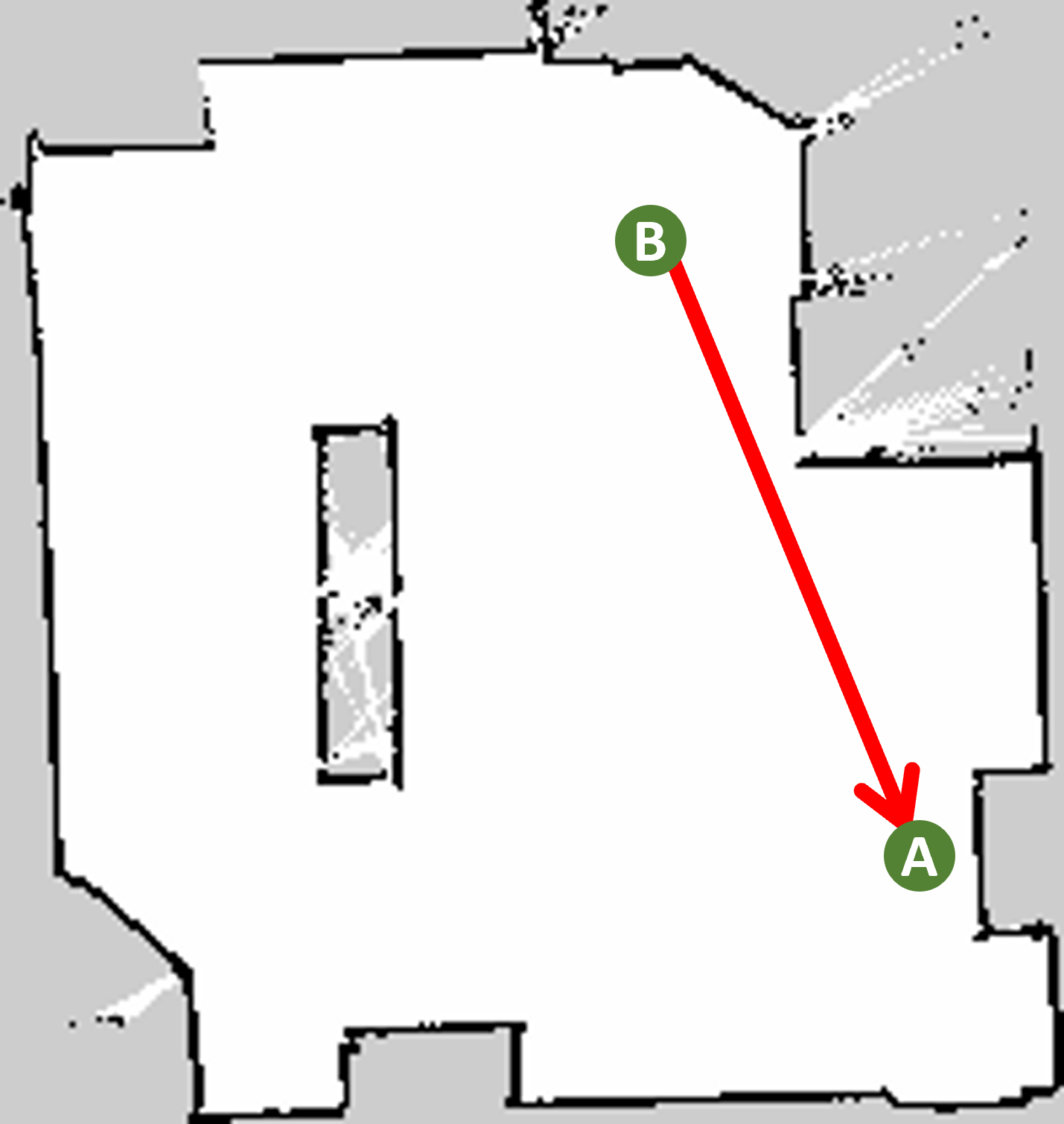}
\caption{Final map with the new path returning to starting point (A) after the intrusion is remedied.}

\label{fig:map3}
\end{figure}

In normal robot operation (\texttt{DEFAULT} mode), it navigates through the scenario from point A to point B and returns to point A (left part of Figure \ref{fig:map1}), crossing the high-security zone. Once started navigating, at point C (central part of Figure \ref{fig:map2}), an intrusion is detected by the emulated RIPS and the system changes to \texttt{ALERT} mode, activating the keep-out zone. During this navigation phase, a person crosses in front of the robot to check that camera is disconnected and to avoid mobile obstacles. After that, the \texttt{COMPROMISED} mode is activated in point D (figure \ref{fig:map2}), and the system stops some devices (twist, camera, laser). Later the robot will return to \texttt{ALERT} mode and continue to point B, where the mode will change to \texttt{DEFAULT}. In the last phase (right part of Figure \ref{fig:map3}), the robot returns to point A following the shortest path.

%\begin{figure}[!htb]
%\minipage{0.23\textwidth}
%  \includegraphics[width=1\linewidth]{map1.png}
%  \caption{Initial map with start (A) and goal (B) points, and first calculated path}\label{fig:map1}
%\endminipage\hfill
%\minipage{0.23\textwidth}
%  \includegraphics[width=1\linewidth]{map2.png}
%  \caption{Intrusion points (C and D), and new path}\label{fig:map2}
%\endminipage\hfill
%\minipage{0.4\textwidth}%
%  \includegraphics[width=0.5\linewidth]{map3.png}
%  \caption{Final map with the new path returning to starting point (A) after the intrusion is remedied}\label{fig:map3}
%\endminipage
%\end{figure}

The robot calculates the path using Nav2~\cite{macenski2020marathon2}, which will be recalculated after the keep zone activation (point C in figure \ref{fig:map2}). We suppose that the area shown in figure \ref{fig:map2} should be excluded from navigation due to contains sensitive information that can be vulnerable to digital privacy. At that moment, the robot will calculate a new path avoiding this area to reach point B and complete the task.

\begin{figure}[h!]
  \centering
  \includegraphics[width=\linewidth]{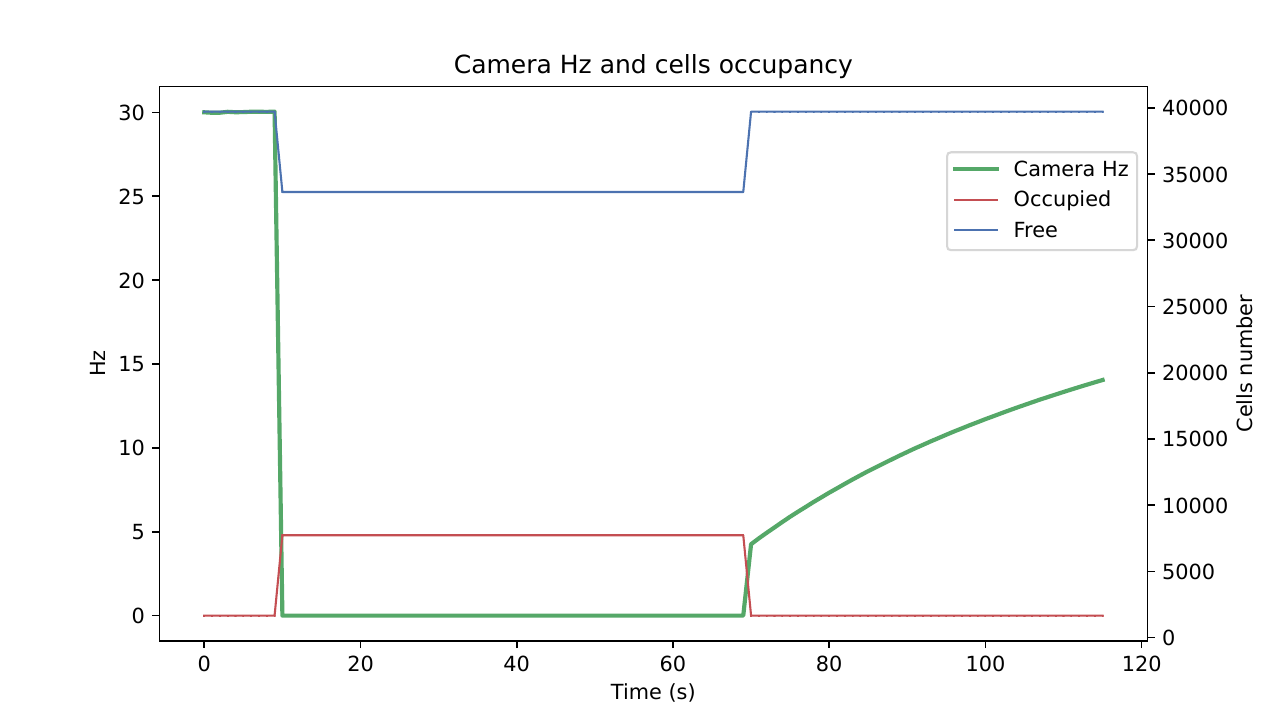}
\caption{Camera Hz and map cells occupancy}
  \label{fig:real_demo}
\end{figure}

Figure \ref{fig:real_demo} shows the behavior of the occupied and free cells of the map, and the hertz at which the graph works after a simulated intrusion.

Figure \ref{fig:real_demo} shows the results for a trial in the scenario with the real robot.
\begin{itemize}
    \item The robot starts in \texttt{DEFAULT} mode.
    \item The robot changes its state when detecting an intrusion at time 12. It goes into a \texttt{ALERT} mode, which disconnects the camera to avoid sending images and activates a security zone through which the robot cannot navigate, maintaining the privacy of the environment and safeguarding a critical area.
    \item The robot continues with its task to reach point B, calculating a new path and disconnecting the camera. The graph shows how the image publication frequency reduces to 0, which means the camera is paused. The number of occupied and free cells decreases after the keep zone activation to avoid the high-security zone.
    \item At time 40, the system goes into a \texttt{COMPROMISED} mode, which keeps the camera disconnected until time 47.
    \item At time 47, the system is changed to \texttt{ALERT} with no apparent changes in behavior mode and finally at time 70 to \texttt{DEFAULT} mode, when the intrusion is mitigated.
    \item The robot is again allowed to navigate through all the available cells of the map and to reach point A will now follow the shortest path due to the keep zone being deactivated. The number of free cells increases to the totality, and the number of occupied cells corresponding to those of the initial map decreases. In addition, when the camera is activated, it starts publishing images, recovering gradually\footnote{After activation, there is a jump, which gradually increases the frequency. DDS (Data Distribution Service for Real-time Systems) is primarily responsible for this problem because when a UDP packet does not contain at least one IP fragment, the rest of the received fragments fill up the kernel buffer. Linux kernels time out after 30 seconds of trying to recombine packet fragments. A full kernel buffer (default size is 256KB) prevents new fragments from coming in at this point, so the connection seems to "hang" for a long time. This issue is generic across all DDS vendors, so the solutions involve adjusting kernel parameters.} the maximum available hertz.
\end{itemize}

\begin{figure*}[h!]
  \centering
  \includegraphics[height=0.28\linewidth]{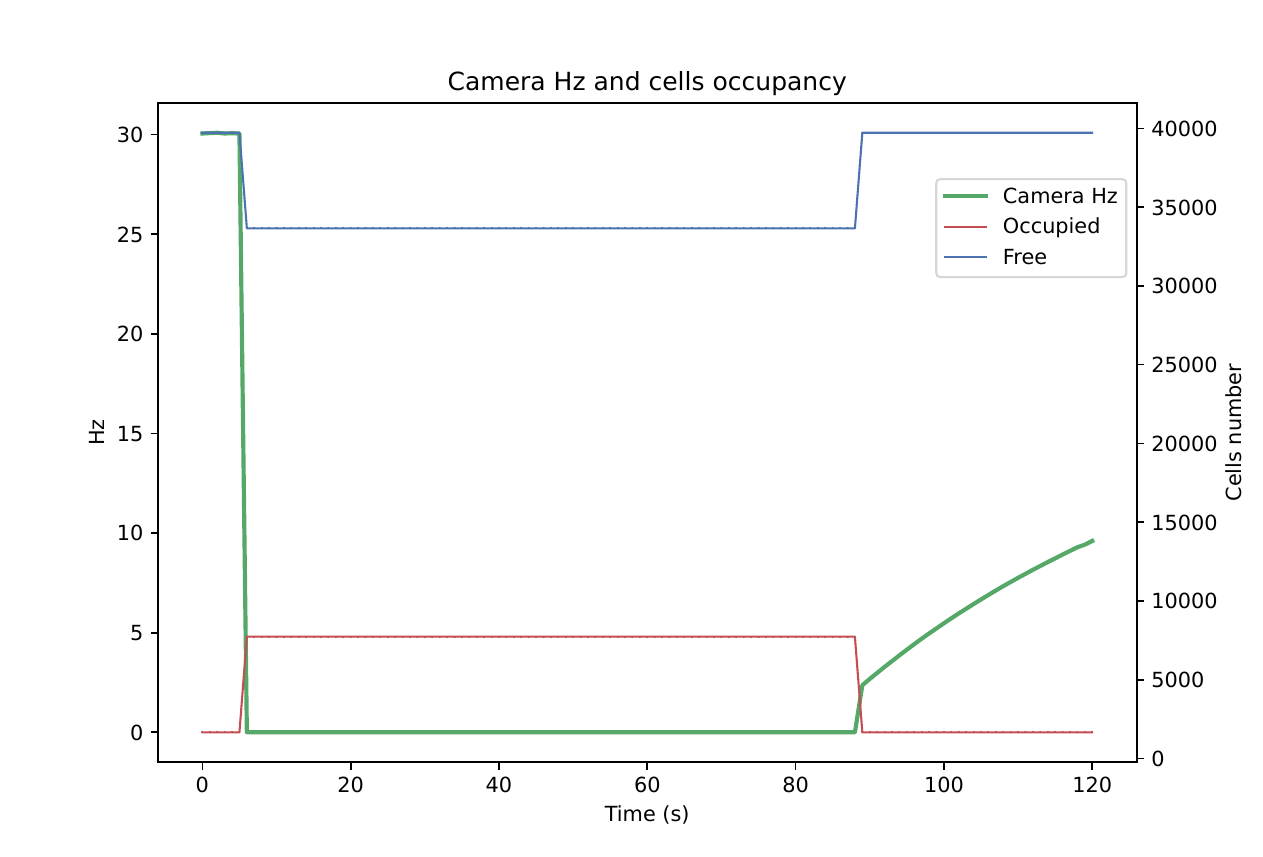}
  \includegraphics[height=0.28\linewidth]{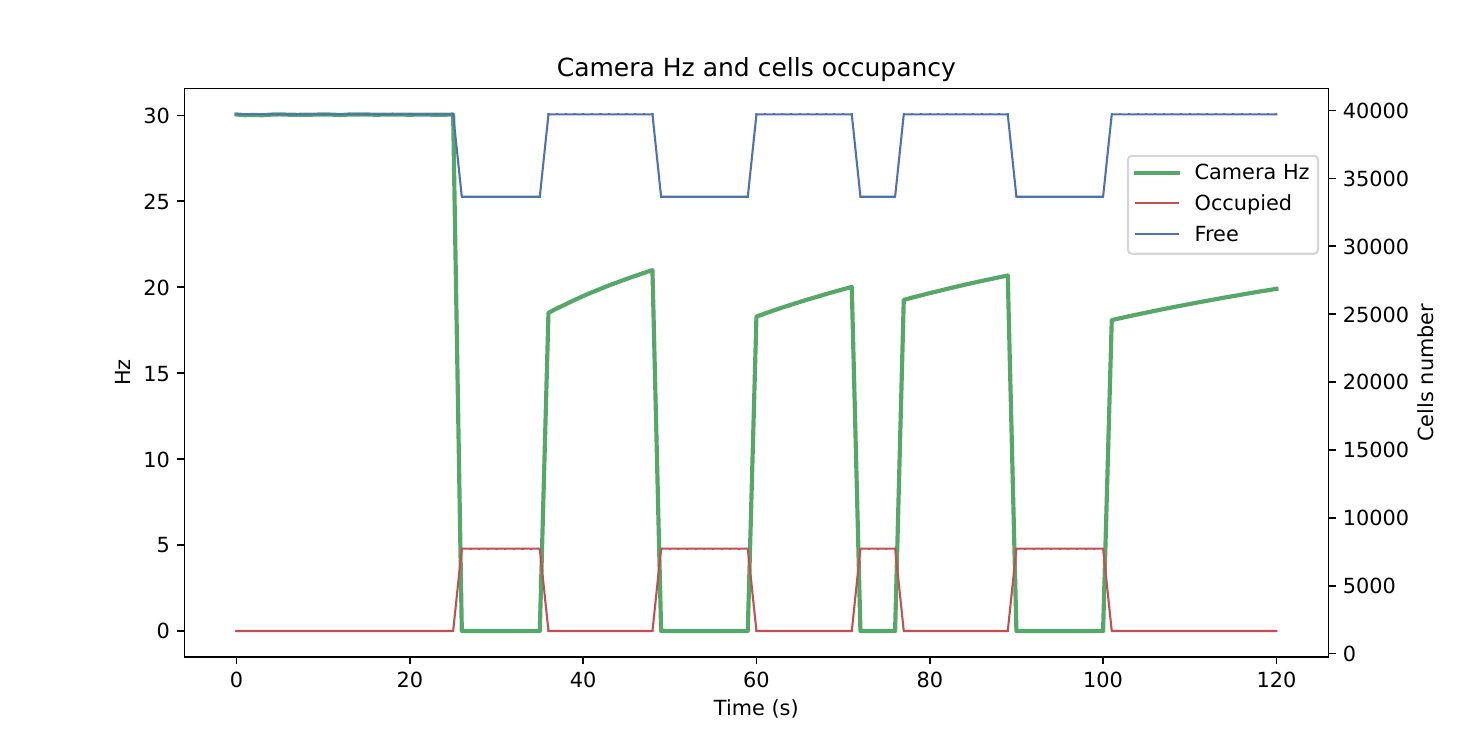}
  \includegraphics[height=0.27\linewidth]{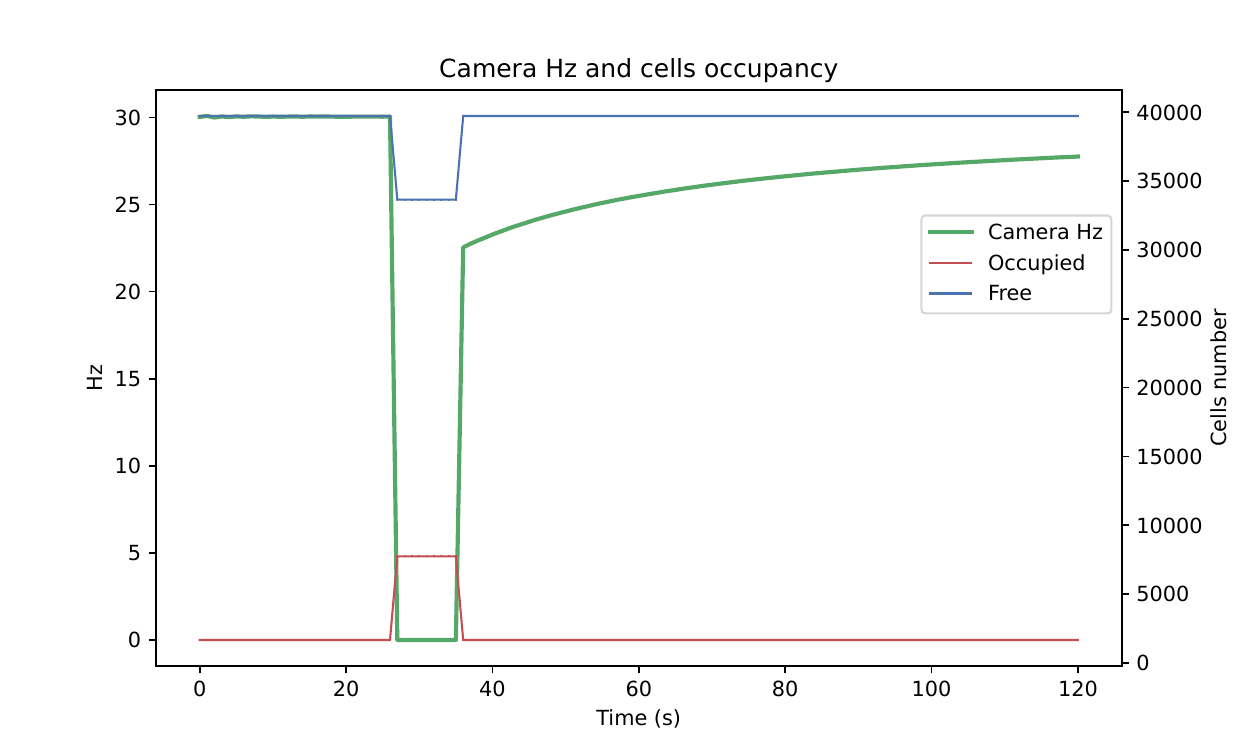}
  \includegraphics[height=0.27\linewidth]{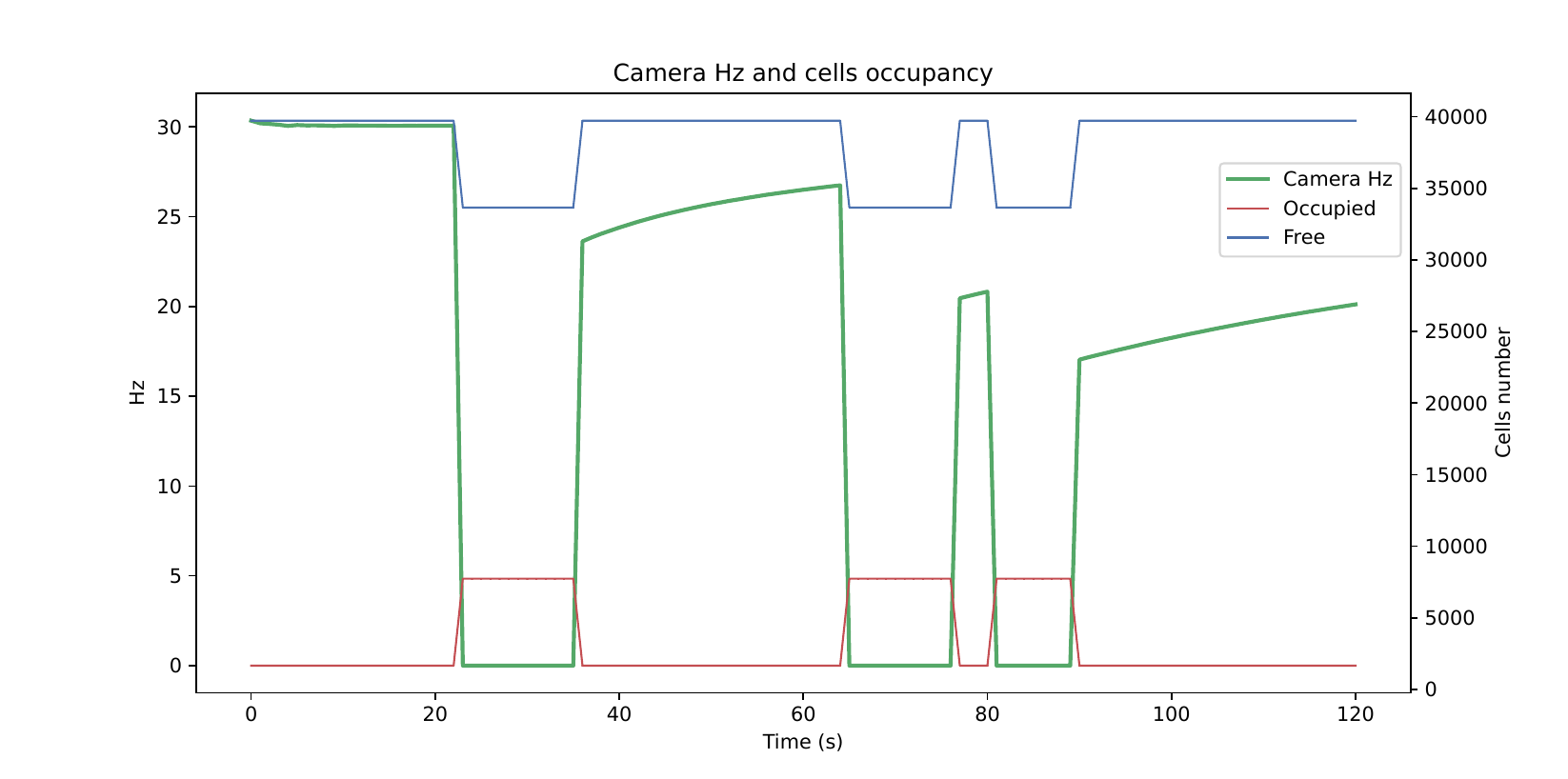}
\caption{amera Hz and cells occupancy. (upper left) long camera deactivation. (upper right)  camera deactivation multiple. (bottom left) short camera deactivation. (bottom right)  camera deactivation multiple with a long activation.}
\label{fig:others}
\end{figure*}

Figure \ref{fig:others} shows another trials with similar behavior. The navegable area is reduced while the threat exists, and the camera stop producing images.

The conclusions of these experiment are that our system correctly reacts to the changes in the security level detected by RIPS. It reconfigures the system as soon as an intrusion is detected, disabling sensible sensors and forbidding the robot navigating in restricted areas. The system also reacts disabling the counter measures when the threat has been mitigated.

%\begin{figure*}
%\centering
%\subfloat[]{\label{fig:test2}\includegraphics[scale=.35]{test2_one.pdf}}
%\subfloat[]{\label{fig:test5_m}\includegraphics[scale=.35]{test5_multi.pdf}}

%\subfloat[]{\label{fig:test5}\includegraphics[scale=.35]{test5_one.pdf}}
%\subfloat[]{\label{fig:test3}\includegraphics[scale=.35]{test3_multi.pdf}}
%\caption{Camera Hz and cells occupancy. (a) long camera deactivation. %(b)  camera deactivation multiple. (c) short camera deactivation. (b)  camera deactivation multiple with a long activation}

%\end{figure*}

\section{Conclusions}
\label{sec:conclusions}

In this work, we have contributed to the state of the art by combining safety and security
through modes of operation applicable to the preservation of privacy and integrity of the environment
during an operation of a social, mobile robot controlled by a cognitive architecture. Throughout this work,
we have described the cognitive architecture, the operating modes system, and one of our major contributions,
the design of a robotic intrusion prevention system (RIPS). We have emphasized technical details to make our
system reproducible. This system can detect and mitigate intrusions in cognitive, social robots, and other
cyber-physical systems.

Our contributions let to separate the operational logic from the contingency management logic by using
a preconfigured set of modes that preserves aspect of safety like privacy or preservation of sensible information.
This is carried out by providing a simple but effective collaboration between RIPS and System Modes, convenient
for real robot applications.

To experimentally validate our contribution, we have planned a scenario where a robot compromised by an intrusion should not use its camera to avoid data robot, nor should it navigate through high-security zones. Experiments have shown that the system reacts correctly when intrusions are detected during the robot's operation.

Future work includes completing a fully functional RIPS prototype to address all kinds of situations through a scripting language that will provide the system with flexibility in its configuration (including a ROS 2 monitor to intercept communications and computation graph changes, and a custom programming language to specify the rules and actions),
creating ROS 2 packages to distribute our prototypes,
and integrating the RIPS with a tamper-evident logging system~\cite{guardiola2022sealfsv2}
and an explainability subsystem.

 \bibliographystyle{elsarticle-num}
 \bibliography{paper}





% \end{thebibliography}
\end{document}